\documentclass[11pt]{article}

\usepackage[preprint]{acl}

\usepackage{times}
\usepackage{latexsym}
\usepackage{booktabs}
\usepackage{fontawesome5}
\usepackage{float}
\usepackage[T1]{fontenc}
\usepackage[utf8]{inputenc}
\usepackage{multirow}
\usepackage{microtype}
\usepackage{inconsolata}
\usepackage{graphicx}
\usepackage{colortbl}

\definecolor{agentblue}{rgb}{0.859, 0.918, 0.996}
\definecolor{deltagreen}{rgb}{0.863, 0.988, 0.906}

\title{AgentFinVQA: A Deployable Multi-Agent Pipeline for Auditable Financial Chart QA}

\author{
  Aravind Narayanan \quad Shaina Raza \\
  Vector Institute \\
  \texttt{\{aravind.narayanan,shaina.raza\}@vectorinstitute.ai}
}

\begin{document}
\maketitle

\begin{abstract}
Financial chart question answering in regulated settings demands more than accuracy: practitioners must know which answers to trust before acting on them, and many institutions cannot send client data to external model providers. Yet existing chart-QA agents are accuracy-focused and opaque, and most assume proprietary API access; to our knowledge, none combines auditability with on-premise deployability without significant accuracy compromise. We present \textbf{AgentFinVQA}, a multi-agent pipeline that decomposes each query into planning, OCR, legend grounding, visual inspection, and verification, recording every step in a traceable Model Evaluation Packet (MEP) per sample. On FinMME, AgentFinVQA improves $+7.68$~pp over a primary-backbone matched zero-shot baseline with a proprietary backbone (Gemini-3 Flash; 71.24\% vs.\ 63.56\%, McNemar $p \approx 1.1\times10^{-16}$), and $+4.84$~pp with open-weights Qwen3.6-27B-FP8 served locally. The verifier's verdict also serves as a useful confidence signal (68.2\% vs.\ 55.6\% exact accuracy on confirmed vs.\ revised answers), enabling human-in-the-loop review routing. Error analysis shows that question misunderstanding, legend confusion and extraction error account for nearly two-thirds of failures and are the categories least detected by the verifier, identifying clear directions for future work. Together these results show that auditable, on-premise financial chart QA is practical and that the open-weights system keeps most of the accuracy gains while enabling full data residency.
We release our code to support reproducible evaluation. \faGithub~\href{https://github.com/VectorInstitute/AgentFinVQA/}{Project Code} 
\end{abstract}

\section{Introduction}
Financial charts are a primary medium for communicating economic data, and automated question answering over such charts could substantially reduce the analytical burden on practitioners \cite{shu2025finchartbenchbenchmarkingfinancialchart}. However, deployment in financial settings demands more than raw accuracy. Recent work shows that such systems exhibit significant hallucination rates on financial tasks, posing direct operational and regulatory risk to practitioners~\cite{zhang2025faith}. A system that is frequently correct but unpredictably wrong is difficult to trust, and one that gives no visibility into its reasoning is impossible to audit. 

\begin{figure*}[t]
  \centering
  \includegraphics[width=0.90\textwidth]{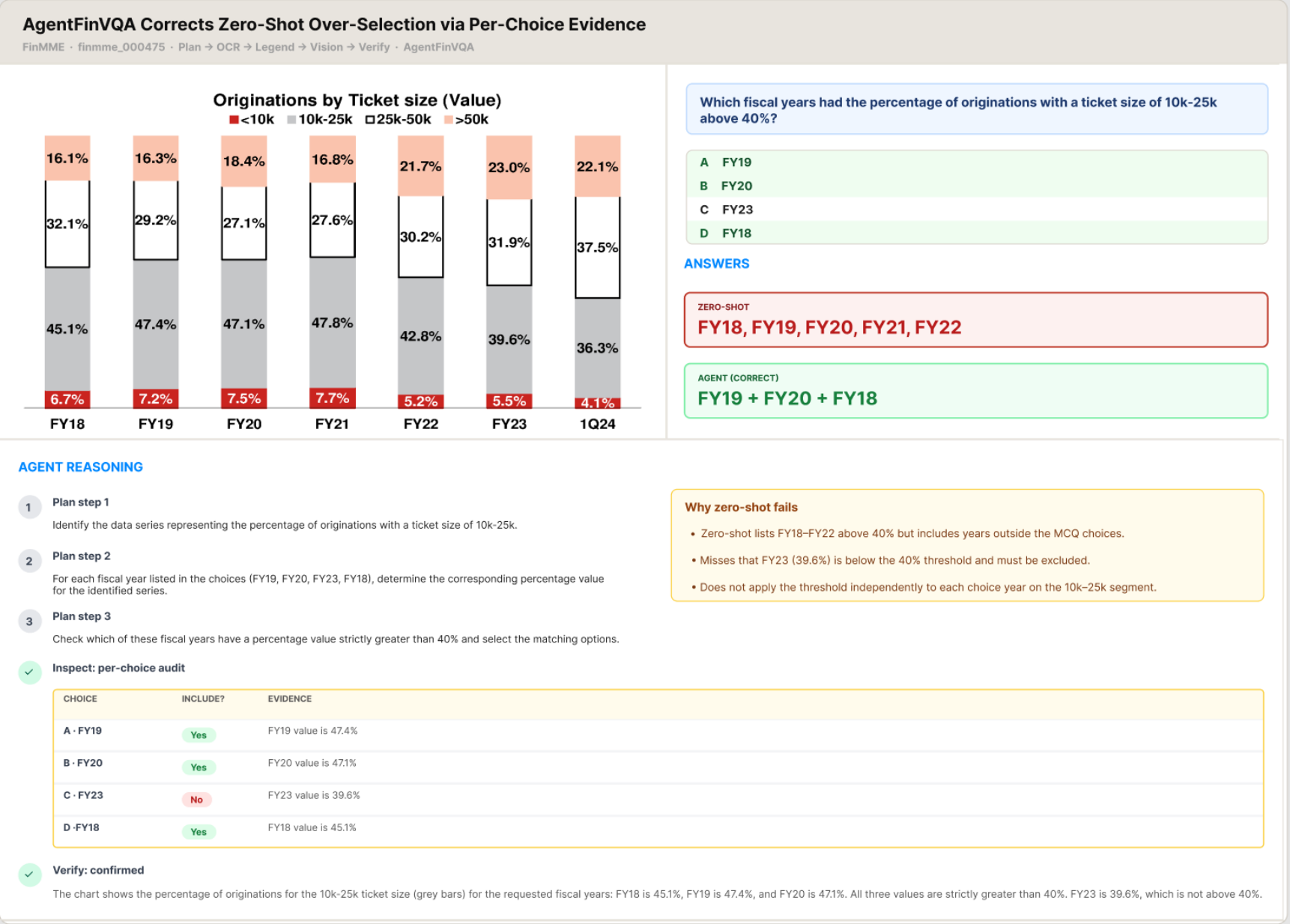}
  \caption{AgentFinVQA corrects zero-shot over-selection on a FinMME stacked ticket-size chart. Zero-shot lists FY18--FY22 for the 10k--25k segment above 40\%, while AgentFinVQA verifies each MCQ option independently, excludes FY23 at 39.6\%, and returns the exact answer: FY18 + FY19 + FY20.}
  \label{fig:teaser}
\end{figure*}

Figure~\ref{fig:teaser} shows a concrete case: zero-shot over-selects fiscal years above the 40\% threshold, while AgentFinVQA checks each MCQ option against the 10k--25k segment values and recovers the exact multi-select answer. A further hard constraint is that many financial institutions cannot send sensitive client documents to external model providers, making accurate local deployment a necessity rather than a preference.

Agentic decomposition, which breaks a chart question into specialised reasoning steps, has improved accuracy on general chart QA \cite{wang2026chartagent, liu-etal-2023-deplot}. The closest systems are agentic chart-QA frameworks such as ChartAgent \cite{kaur2026chartagentmultimodalagentvisually}, which decompose queries into visual subtasks but rely on local computer-vision tools to manipulate the image, and ChartSketcher \cite{huang2026chartsketcher}, which requires a two-stage fine-tune on hundreds of thousands of annotated samples.
To our knowledge, no prior system delivers chart QA that is at once prompting-only (no local segmentation models or task-specific fine-tuning), auditable per answer, and deployable on open weights in-house, the combination that regulated financial settings require. 

In this work we ask: \textit{can a financial chart-QA system be auditable and run entirely in-house without sacrificing accuracy?} To answer this,  we develop \textbf{AgentFinVQA}, a multi-agent pipeline that coordinates a text-only planner, an OCR reader, a legend grounder, a lightweight deterministic colour-area measurement stage, and a vision-and-verify loop, recording each stage in a Model Evaluation Packet (MEP) for full auditability and confidence-based human review routing.
\begin{figure*}[t]
  \centering
  \includegraphics[width=0.9\textwidth]{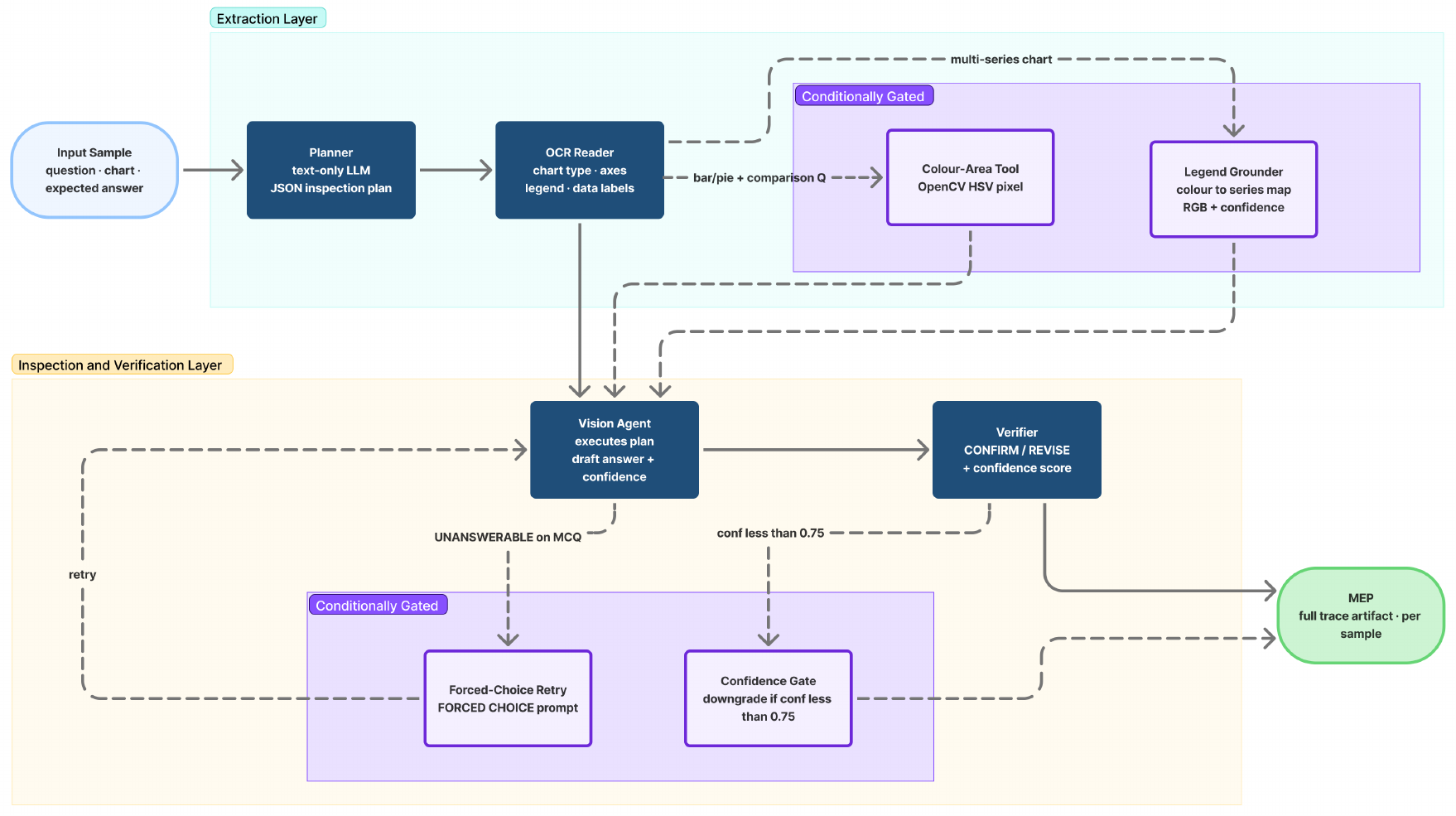}
  \caption{AgentFinVQA pipeline. Input flows left-to-right through five required stages (blue/teal): Planner, OCR Reader, Vision Agent, Verifier, and MEP output. Purple boxes indicate conditionally gated stages: Legend Grounder (multi-series charts), Colour-Area Tool (bar/pie comparison questions), Forced-Choice Retry (MCQ over-refusal), and Confidence Gate (low-confidence verifier revisions).}
  \label{fig:pipeline}
\end{figure*}

On the FinMME benchmark \cite{luo-etal-2025-finmme}, our pipeline with a proprietary backbone (Gemini-3) improves \textbf{+7.68~pp} over a model-matched zero-shot baseline ($p \approx 1.1 \times 10^{-16}$), with the largest gains on MCQ questions (\textbf{+8.1~pp}). The same pipeline with open-weights Qwen3.6-27B-FP8, served locally on a single A100, yields \textbf{+4.84~pp} ($p \approx 3.0 \times 10^{-6}$), confirming the gains do not depend on a proprietary API.

Our contributions are: (1) \textbf{AgentFinVQA}, a multi-agent chart-QA pipeline whose every step is recorded in a traceable MEP, making each answer auditable; (2) evidence that accuracy gains transfer from proprietary to locally-served open-weights models, supporting in-house deployment with modest accuracy trade-off; and (3) a verifier verdict that acts as a confidence signal, letting human reviewers focus on the answers most likely to be wrong.

\section{Related Work}
\label{sec:related}

\paragraph{Chart QA benchmarks.} Early datasets such as FigureQA \cite{kahou2018figureqaannotatedfiguredataset}, DVQA \cite{kafle2018dvqa}, and PlotQA \cite{methani2020plotqa} established the task on synthetic charts, while ChartQA \cite{masry-etal-2022-chartqa} and ChartQA Pro \cite{masry-etal-2025-chartqapro} introduced human-written questions over real charts. FinMME \cite{luo-etal-2025-finmme},  FinChart-Bench \cite{shu2025finchartbenchbenchmarkingfinancialchart} and MME-Finance \cite{gan2024woodpecker} confirm the difficulty of the financial setting but offer no deployable agentic solution, the gap we address.

\paragraph{Decomposition and structured extraction.} DePlot \cite{liu-etal-2023-deplot} converts charts into data tables for one-shot LLM reasoning, separating modality conversion from reasoning; this informs our design, where the OCR Reader and Legend Grounder produce structured text metadata that grounds later stages. Unlike MatCha \cite{liu-etal-2023-matcha} and UniChart \cite{masry-etal-2023-unichart}, which require domain-specific pretraining, our pipeline achieves comparable extraction through prompting alone and works with any vision language model (VLM) backend.

\paragraph{Agentic chart understanding.} ReAct \cite{yao2023react} established the reason-and-act paradigm; our Plan, OCR, Ground, Inspect, Verify pipeline applies it to  chart QA. The closest systems, ChartAgent \cite{kaur2026chartagentmultimodalagentvisually} and a YOLO-based variant \cite{wang2026chartagent}, decompose queries into visual subtasks but physically manipulate the chart image with local computer-vision models. AgentFinVQA instead injects OCR output and legend maps as text, avoiding local segmentation infrastructure. ChartSketcher \cite{huang2026chartsketcher} reaches strong results but requires a two-stage fine-tune on 300K samples with RL; our pipeline is prompting-only and runs on both proprietary and open-weights backends. MAC-SQL \cite{wang-etal-2025-mac} shows the same plan--reason--verify pattern in text-to-SQL.

\paragraph{Verification and deployment.} Prior work reduces hallucination through self-correction or judge-based analysis \cite{pan2026through,NEURIPS2023_91f18a12}; our verifier follows this intuition but runs as a prompting-only inference stage and provides a confidence signal for review routing. We use judge-based analysis only for post-hoc failure taxonomy, while answer accuracy is measured by the rule-based scorer in Section~\ref{sec:setup}. FAITH \cite{zhang2025faith} motivates the need to manage silent failures in financial settings.

 \section{AgentFinVQA Framework}
\label{sec:framework}
\paragraph{Problem setup.}
Given a financial chart image $c$, a natural-language question $q$, and an optional set of MCQ choices $O = \{o_1,\dots,o_k\}$, the task is to produce an answer $a$ together with an auditable trace $\mathcal{M}$ of the reasoning that led to it. For open-ended (standard) questions $a$ is a numeric value or short text string; for single-select MCQ $a \in O$; for multi-select MCQ $a \subseteq O$. An answer is scored by a rule-based scorer that applies numeric tolerance to standard answers and partial credit to MCQ answers, giving a per-sample score in $[0,1]$ (mean answer accuracy); a sample counts as exactly correct when its score is $\geq 0.999$ (exact accuracy). Beyond producing $a$, we impose two deployment constraints that the architecture is built to satisfy while preserving strong answer accuracy: the system must be (i) \emph{prompting-only}, using no task-specific fine-tuning and no local computer-vision models, and (ii) runnable on a self-hosted open-weights backbone, so sensitive charts never leave the institution.

AgentFinVQA is a multi-agent system of specialised agents, a planner, OCR reader, legend grounder, deterministic colour-area tool, vision agent, and verifier, that operate sequentially over $(c, q, O)$ and produce a verified answer alongside a fully traceable \textbf{Model Evaluation Packet (MEP)}. The pipeline executes stages in a fixed order, with gated stages skipped when their trigger conditions are not met: Plan $\rightarrow$ OCR $\rightarrow$ Ground $\rightarrow$ Colour-Area $\rightarrow$ Inspect $\rightarrow$ Verify. The order is not arbitrary: each stage's structured output becomes structured evidence for the next, so cheaper and more reliable text extraction (OCR, legend) constrains the harder visual estimation that follows, and verification runs last against an independent view of the chart. Figure~\ref{fig:pipeline} shows the architecture. Every stage writes its inputs, outputs, tool traces, and timestamps into the per-sample MEP, a portable JSON artifact (see Appendix~\ref{sec:appendix_mep}) enabling reproducible evaluation, post-hoc error attribution, and audit without re-running the pipeline. The formal definitions of all stage abbreviations and terminology used throughout this section are provided in Appendix~\ref{sec:appendix_notation}.

\paragraph{Planner.}
A text-only LLM receives the question and MCQ choices but does not see the chart image. It outputs a structured JSON inspection plan comprising two to three focus points, a question type classification, and an answerability assessment. For MCQ questions, the planner explicitly instructs the vision agent to check \textit{each} choice independently and verify that the selected answer is not contradicted by other data. For multi-select questions, it instructs the agent to compile \textit{all} supported choices rather than stopping at the first match.

\paragraph{OCR Reader.}
A single focused VLM call transcribes all visible text, producing structured metadata: chart type, axis labels and ticks, legend entries, data labels, and annotations (see Appendix~\ref{sec:appendix_mep}, stage field structure). This output serves as structured evidence for visible text in all subsequent stages, preventing the vision agent from misreading labels already reliably extracted.

\paragraph{Legend Grounder.}
A targeted VLM call maps each legend entry to its visual properties: colour description, approximate RGB, line style, and confidence. This legend map is injected into the vision prompt as explicit structured evidence with the instruction not to reassign colours during value extraction. A compliance check verifies that the vision agent's explanation references at least one legend label by name; if not, the vision call is retried. Compliance retries fired on 13.2\% of MEPs, confirming the check is actively catching non-compliance. The stage is gated on chart type and legend size.
\begin{table*}[t]
  \centering
  \small
  \begin{tabular}{llcccc}
    \toprule
    \textbf{System} & \textbf{Backbone} &
    \textbf{Mean Acc.} & \textbf{Exact} &
    \textbf{MCQ Mean} & \textbf{Open Question Mean} \\
    \midrule
    Zero-shot & Gemini-3 Flash
      & 63.56\% & 56.40\% & 64.4\% & 54.0\% \\
    \rowcolor{agentblue}
    AgentFinVQA & Gemini-3 Flash
      & \textbf{71.24\%} & \textbf{65.12\%}
      & \textbf{72.5\%} & \textbf{57.0\%} \\
    \rowcolor{deltagreen}
    $\Delta$ & & \textbf{+7.68 pp} & \textbf{+8.72 pp}
      & \textbf{+8.1 pp} & \textbf{+3.0 pp} \\
    \cmidrule(lr){1-6}
    Zero-shot & Qwen3.6-27B-FP8
      & 61.68\% & 53.52\% & 62.8\% & 49.0\% \\
    \rowcolor{agentblue}
    AgentFinVQA & Qwen3.6-27B-FP8
      & \textbf{66.52\%} & \textbf{60.24\%}
      & \textbf{68.1\%} & \textbf{48.0\%} \\
    \rowcolor{deltagreen}
    $\Delta$ & & \textbf{+4.84 pp} & \textbf{+6.72 pp}
      & \textbf{+5.3 pp} & $-$1.0 pp$^\dagger$ \\
    \bottomrule
  \end{tabular}
  \caption{AgentFinVQA vs.\ model-matched zero-shot baseline on FinMME.
    McNemar: Gemini-3 $p \approx 1.1 \times 10^{-16}$;
    Qwen3.6-27B-FP8 $p \approx 3.0 \times 10^{-6}$.
    $^\dagger$Standard $\Delta$ for Qwen within noise (CI\,$\approx$\,$\pm$10\,pp).}
  \label{tab:main}
\end{table*}

\paragraph{Colour-Area Tool.}
To address perceptual estimation errors on stacked bar and pie charts, we introduce a \textbf{deterministic pixel-counting stage} between legend grounding and vision. Given the legend RGB map produced by the previous stage, the tool applies per-series HSV colour masks ($\pm$10 hue, $\pm$40 saturation/value) to count matched pixels per legend entry, producing a dominant-label hint injected into the vision prompt. Unlike the local computer-vision models used by prior
agentic systems~\cite{kaur2026chartagentmultimodalagentvisually}, this stage requires no GPU, no trained model, and no segmentation infrastructure; it is a lightweight rule-based computation that runs anywhere Python and OpenCV are available.

The stage is gated conservatively: it fires only when chart type is bar, pie, or donut; the legend has $>1$ entry; the question contains a comparison keyword (\textit{largest, smallest, most, least}, etc.); and no colour ambiguity is detected between legend entries (pairwise HSV hue distance $>15$). Suppression flags and the full pixel breakdown are stored in the \texttt{MEPColorArea} field for post-hoc audit (Appendix~\ref{sec:appendix_mep}).

\paragraph{Vision Agent.}
A CrewAI-orchestrated agent \cite{crewai} executes the inspection plan with the chart image, OCR metadata, legend map, and colour-area hint (when available). It produces a draft answer, explanation, and per-choice confidence scores. Three prompt paths handle single-select MCQ, multi-select MCQ, and open-ended questions. If the vision agent returns UNANSWERABLE on an MCQ question, a forced-choice retry re-runs with an explicit instruction to select the most plausible option, reducing the UNANSWERABLE rate.

\paragraph{Verifier.}
A second independent VLM call audits the draft answer against the chart image. It receives the draft answer, explanation, per-choice analysis, MCQ choices, analyst caption, and the sample's related sentences. Trace analysis showed that many confirmed-wrong cases contradicted information explicitly stated in this field. The verifier produces a \texttt{CONFIRM} or \texttt{REVISE} verdict and a self-reported confidence score, both recorded in the per-sample MEP (Appendix~\ref{sec:appendix_mep}). A \textbf{confidence gate} downgrades revisions with confidence $<0.75$ to confirmations, preventing the verifier from overriding high-quality vision answers with uncertain revisions. Note that the verifier is distinct from the evaluation judge: the verifier is a pipeline stage that runs on every sample at inference time, whereas the judge is a separate model used only for post-hoc failure categorisation in Section~\ref{error_analysis}.

\paragraph{Backend flexibility.}
All LLM stages route through a configurable backend abstraction supporting the Gemini API, OpenAI-compatible endpoints, and locally-served models via vLLM, allowing different stages to use different models — lightweight for structured extraction (OCR, legend grounding) and more capable for reasoning-intensive stages (planning, vision, verification). For the open-weights evaluation we serve \textbf{Qwen3.6-27B-FP8}~\cite{qwen3.6-27b} on a single A100-80G with no prompt modification, satisfying on-premise data-residency requirements for chart inference without any external API dependency.

\begin{table*}[t]
  \centering
  \small
  \resizebox{0.75\textwidth}{!}{
  \begin{tabular}{p{0.24\textwidth}p{0.42\textwidth}p{0.2\textwidth}}
    \toprule
    \textbf{Component} & \textbf{Contribution} & \textbf{Effect} \\
    \midrule
    MCQ-aware planning + forced-choice retry
      & Eliminates over-refusal on MCQ; forces a best-guess when the model would otherwise abstain
      & Near-elimination of MCQ abstentions \\
    \addlinespace
    Multi-select MCQ support
      & Dedicated prompts for select-all-that-apply questions
      & +23.3 pp on \texttt{multiple\_choice} \\
    \addlinespace
    Verifier with confidence gate
      & Second VLM pass with gated revision; confirmed answers are reliably more accurate than revised ones
      & 68.2\% (conf.)\ vs.\ 55.6\% (rev.) \\
    \addlinespace
    Colour-area tool
      & Deterministic pixel-counting for stacked bar/pie comparisons; grounding stored in MEP audit trace
      & 5\% activation \\
    \bottomrule
  \end{tabular}
  }
  \caption{Component contributions to the AgentFinVQA pipeline. Multi-select MCQ support is the largest isolated accuracy gain. The colour-area tool's primary value is deterministic grounding and audit traceability rather than aggregate accuracy lift.}
  \label{tab:components}
\end{table*}

\section{Results}
\subsection{Experimental Setup}
\label{sec:setup}

\textbf{Dataset.} We evaluate on \textbf{FinMME}~\cite{luo-etal-2025-finmme}, a financial VQA benchmark of $\sim$11{,}000 samples spanning bar, line, pie, stacked bar, and combination charts across single-select MCQ, multi-select MCQ, and open-ended standard question formats.

\textbf{Models.} We compare AgentFinVQA against primary-backbone matched zero-shot baselines, a single structured VLM call requesting
JSON output: \textbf{Gemini-3 Flash}~\cite{google2025gemini3flash} and \textbf{Qwen3.6-27B-FP8} for the open-weights configuration. All systems receive the same auxiliary analyst caption and related-sentence fields when available, so these inputs are controlled across comparisons. In the Gemini-based configuration, Gemini-3 Flash handles planning, vision, and verification; OCR and legend grounding use Gemini-2.5 Flash Lite, which suffices for structured text extraction at lower cost. In the Qwen-based configuration, Qwen3.6-27B-FP8 runs all inference stages via vLLM on a single A100-80G; Gemini Batch API is used only for post-hoc error analysis.

\textbf{Metrics.} Accuracy is measured by a rule-based scorer with numeric tolerance for standard answers and partial credit for MCQ (\textbf{mean answer accuracy}), and by the fraction of samples scoring $\geq 0.999$ (\textbf{exact accuracy}). Statistical significance uses the paired \textbf{McNemar test} on binary strict
correctness.

\subsection{Performance Evaluation}
\label{sec:results}
Our first experiment tests the core claim: \textit{does agentic decomposition beat a model-matched zero-shot baseline, and does the gain hold across both a proprietary and an open-weights backbone?} Table~\ref{tab:main} presents the comparison.

The agent improves over zero-shot on both backbones, confirming the gains are not artefacts of a specific proprietary model. MCQ questions drive the aggregate on both backbones (+8.1 pp Gemini, +5.3 pp Qwen), while standard open-ended questions show a smaller gain for Gemini (+3.0 pp) and no reliable gain for Qwen ($-$1.0 pp, wide CI). The smaller overall Qwen gain (4.84 pp vs.\ 7.68 pp) is partly attributable to verifier over-revision: the Qwen verifier modifies 41\% of vision-agent answers (vs.\ 17\% for Gemini), and revised answers score substantially lower than confirmed ones, suggesting the verifier frequently overrides correct vision outputs.

\subsection{Ablations}
\label{sec:ablation}
Our second experiment isolates the contribution of each major design decision, drawn from the most cleanly isolated comparisons in our development data (Table~\ref{tab:components}). Development runs were conducted on a 25\% sample of FinMME ($n \approx 2{,}775$) due to API cost constraints.

The most defensible isolated gain is \textbf{multi-select MCQ support}: adding separate planner, vision, and verifier prompt paths for select-all-that-apply questions raised \texttt{multiple\_choice} accuracy from 30.2\% to 53.5\% (\textbf{+23.3 pp}), the largest single-component contribution in the pipeline. The full development progression is in Appendix~\ref{sec:appendix_progression}.

\subsection{Error Analysis and Uncertainty} \label{error_analysis}

\begin{figure}[t]
  \centering
  \includegraphics[width=\columnwidth]{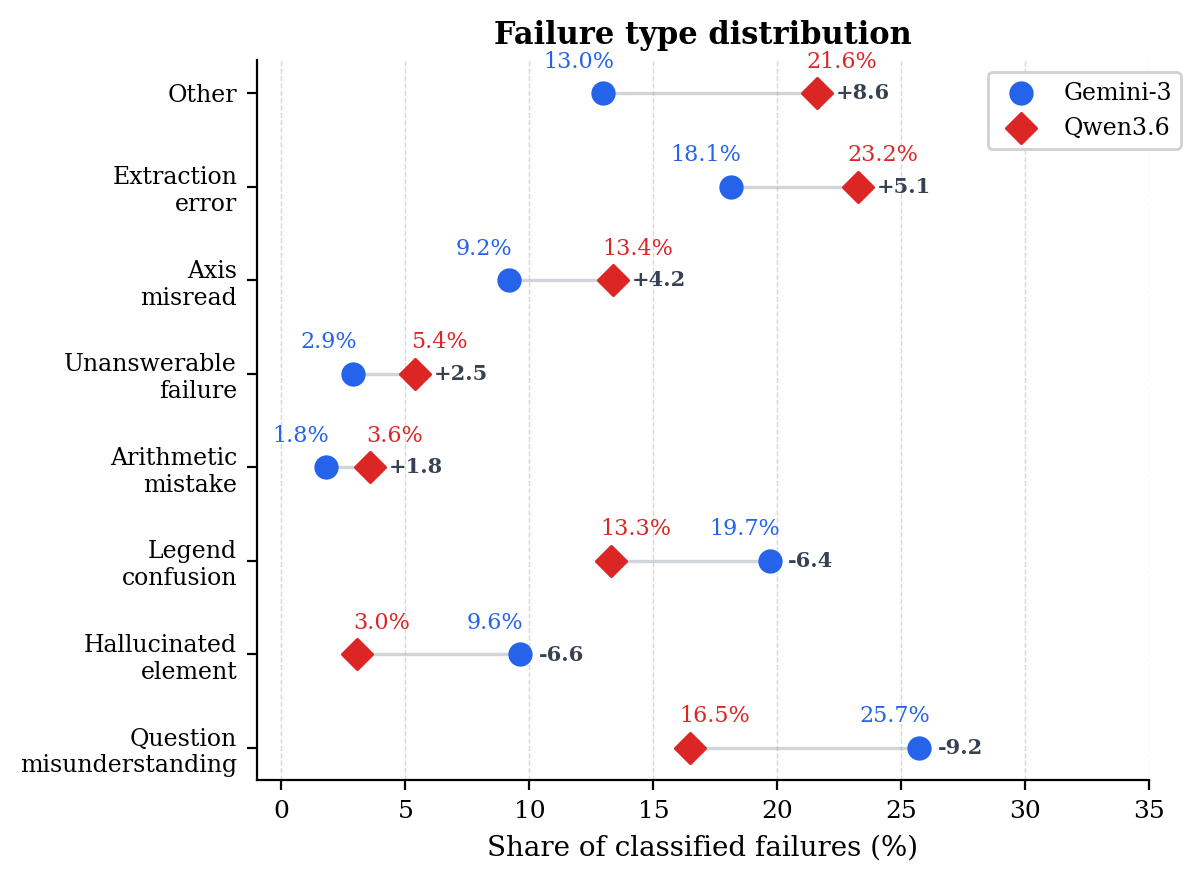}
  \caption{Failure type distribution across backbones (deltas: Qwen3.6 $-$ Gemini-3). The proprietary backbone fails more on question misunderstanding and hallucinated elements; the open-weights backbone shows higher extraction error and axis misread
  rates.}
  \label{fig:failure_distribution}
\end{figure}

We classify all the incorrect outputs from the best run using a judge-generated taxonomy (Figure~\ref{fig:failure_distribution}). Three major categories account for nearly two-thirds of the Gemini-3 failures. \textbf{Question misunderstanding}, where the agent answers a related but different question (typically inverting a superlative or misreading a trend direction); \textbf{legend confusion}, referring to the incorrect series-to-colour mapping, most common on multi-series line or stacked complex charts; and \textbf{extraction error} for wrongly reading the numeric value despite identifying the correct chart element. The remaining failures are split across hallucinated elements, axis misreads, and other minor categories.

\begin{figure}[t]
  \centering
  \includegraphics[width=\columnwidth]{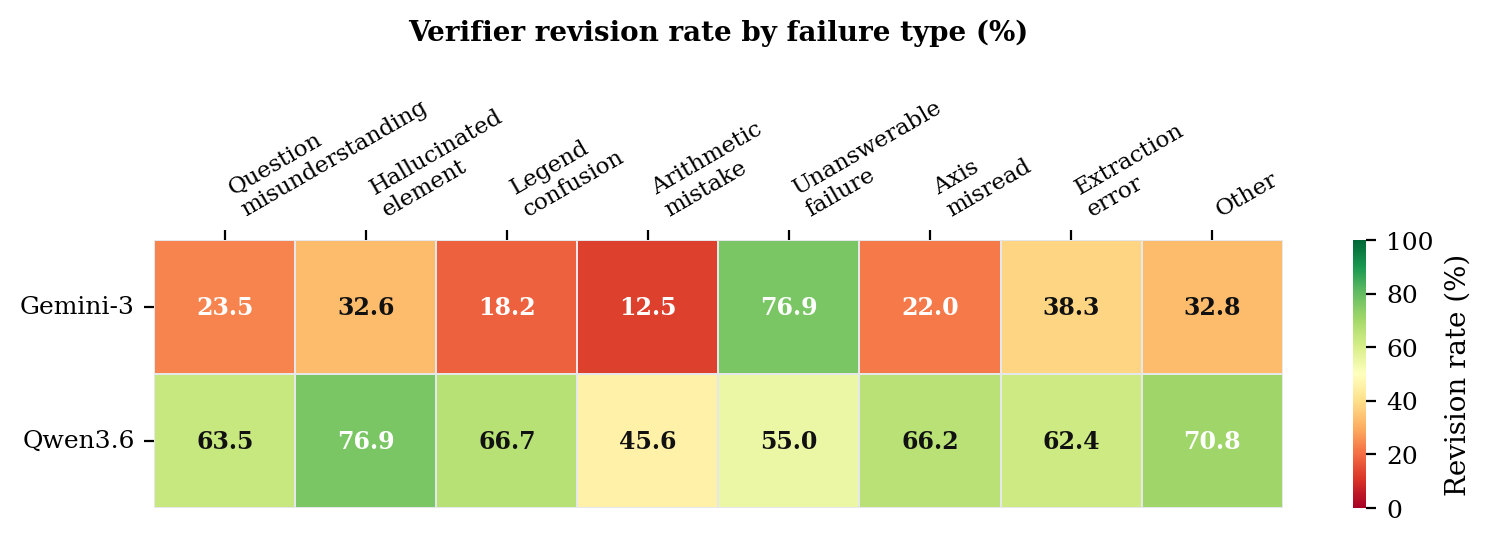}
  \caption{Verifier revision rate per failure category. Green cells indicate failure modes the verifier frequently flags for revision; red cells indicate under-detection. Qwen3.6 consistently shows higher revision rates across all categories, reflecting greater verifier intervention on a weaker backbone.}
  \label{fig:failure_revision}
\end{figure}
\paragraph{Verifier as a routing signal.}
Figure~\ref{fig:failure_revision} shows the verifier revision rate per failure category. A practitioner can implement a simple routing rule: \texttt{revised} answers are prioritized for analyst review, while \texttt{confirmed} answers are treated as lower-priority review items. This concentrates human effort on the $\sim$19\% of outputs most likely to be wrong, while the remaining $\sim$81\% achieve 68\% exact accuracy, offering a pragmatic accuracy-auditability tradeoff for financial deployment.

\section{Conclusion}

We presented AgentFinVQA, a multi-agent pipeline for financial chart question answering that decomposes each query into planning, OCR, legend grounding, visual inspection, and verification, recording every step in a traceable Model Evaluation Packet. On FinMME it improves +7.68\,pp over a model-matched zero-shot baseline with a proprietary backbone and +4.84\,pp with an open-weights model served locally, confirming that the gains do not depend on a proprietary API. The verifier's verdict serves as a review-routing signal, letting analysts focus on the outputs most likely to require correction. Error analysis reveals that question misunderstanding, legend confusion and extraction error account for nearly two-thirds of remaining Gemini-3 failures, precisely the categories the verifier is least effective at self-detecting. Together these results show that auditable, on-premise financial chart QA is practical while preserving the accuracy gains of an agentic approach.
\section*{Limitations}

We acknowledge some limitations of our evaluation. First, we evaluate only on FinMME; while it is large and domain-diverse, we have not tested cross-dataset generalization to established financial chart benchmarks such as FinChart-Bench or MME-Finance, so transfer to other annotation styles and chart distributions remains open. Second, the open-weights configuration recovers most of the MCQ gain but shows no reliable gain on open-ended standard questions ($-$1.0\,pp, within noise), indicating that on-premise deployment with open-weight models still carries a modest accuracy cost on the hardest question type. Third, the colour-area tool activates on only 5\% of dataset and is suggestive rather than conclusive at this distribution; its contribution is architectural (deterministic grounding and audit trace) rather than aggregate accuracy, and its benefit may be larger on chart distributions richer in stacked-bar and pie comparisons. Finally, the verifier's confidence is self-reported rather than calibrated, and although we show that its \texttt{CONFIRM}/\texttt{REVISE} verdict separates answer quality, we do not yet evaluate the proposed human-in-the-loop routing with real analysts; quantifying reviewer time saved is left to future work.

\section*{Acknowledgments}

Resources used in preparing this research were provided, in part, by the Province of Ontario and the Government of Canada through CIFAR, as well as companies sponsoring the Vector Institute (\url{http://www.vectorinstitute.ai/#partners}).

This research was funded by the European Union’s Horizon Europe research and innovation programme under the AIXPERT project (Grant Agreement No. 101214389), which aims to develop an agentic, multi-layered, GenAI-powered framework for creating explainable, accountable, and transparent AI systems.

\bibliography{custom}

\appendix

\section{Notation and Stage Definitions}
\label{sec:appendix_notation}

Table~\ref{tab:notation} defines the key terms and stage abbreviations used throughout this paper.

\begin{table}[h]
  \centering
  \small
  \begin{tabular}{p{0.22\columnwidth}p{0.68\columnwidth}}
    \toprule
    \textbf{Term} & \textbf{Definition} \\
    \midrule
    VQA      & Visual Question Answering: the task of answering a natural language question about an image. \\
    VLM      & Vision-Language Model: a model that jointly processes image and text inputs. \\
    MCQ      & Multiple-Choice Question: a question with a fixed set of answer options. \\
    MEP      & Model Evaluation Packet: a per-sample JSON artifact recording all stage inputs, outputs, tool traces, and timestamps for a single pipeline run. \\
    PLAN     & Planner stage: a text-only LLM that produces a structured JSON inspection plan without seeing the chart image. \\
    OCR      & Optical Character Recognition stage: a focused VLM call that transcribes all visible text in the chart into structured metadata. \\
    GROUND   & Legend Grounding stage: a targeted VLM call that maps each legend entry to its visual properties (colour, RGB, line style, confidence). \\
    COLOUR-AREA & Colour-Area Tool stage: a deterministic pixel-counting stage that applies HSV color masks per legend entry and injects a dominant-label hint into the vision prompt. \\
    INSPECT  & Vision Agent stage: a CrewAI-orchestrated agent that executes the inspection plan against the chart image and produces a draft answer. \\
    VERIFY   & Verifier stage: an independent VLM call that audits the draft answer, producing a \texttt{CONFIRM} or \texttt{REVISE} verdict with a confidence score. \\
    CONF. GATE & Confidence Gate: a rule that downgrades a \texttt{REVISE} verdict to \texttt{CONFIRM} when verifier confidence $< 0.75$. \\
    ZS       & Zero-shot: a single VLM call with a structured prompt; no pipeline stages. Used as the model-matched baseline. \\
    pp       & Percentage points: absolute difference between two percentages. \\
    \bottomrule
  \end{tabular}
  \caption{Notation and stage definitions used throughout the paper.}
  \label{tab:notation}
\end{table}

\section{MEP Schema and Example}
\label{sec:appendix_mep}

Each pipeline run produces one \textbf{Model Evaluation Packet (MEP)}: a portable JSON artifact that records every stage's inputs, outputs, parsed results, tool traces, and wall-clock timestamps. Figure~\ref{fig:pipeline} in the main paper points to the MEP as the pipeline's audit output; this appendix shows its structure.

\paragraph{Top-level fields.} \label{par:mep-toplevel}
The MEP contains the following top-level keys:

{\small
\begin{verbatim}
{
  "schema_version":   "mep.v1",
  "run_id":           "1748647a-...",
  "sample_id":        "finmme_000006",
  "config": {
    "planner_model":  "gemini-3-flash-preview",
    "vision_model":   "gemini-3-flash-preview",
    "judge_backend":  "gemini"
  },
  "plan":             { ... },
  "ocr":              { ... },
  "legend_grounding": { ... },
  "color_area":       null,
  "vision":           { ... },
  "verifier":         { ... },
  "answer":           "C",
  "answer_accuracy":  1.0,
  "verifier_verdict": "confirmed",
  "timestamps":       { "start": 
                            "2026-05-12T19:48:34Z",
                        "end":   
                            "2026-05-12T19:49:18Z" },
  "errors":           [],
  "lf_trace_id":      "2d62cdb2-..."
}
\end{verbatim}
}

\paragraph{Stage field structure.} \label{par:mep-stage}
Every stage field (e.g.\ \texttt{ocr}, \texttt{vision}) follows a common schema:

{\small
\begin{verbatim}
{
  "chart_type":  "bar",
  "x_axis":      { "ticks": ["FY19","FY20","FY21",
                              "FY24F","FY25F","FY26E"] },
  "legend":      ["Employee cost (INR Mn) (LHS)",
                  "Rental cost (INR Mn) (LHS)",
                  "Overheads as a % of sales (RHS)"],
  "data_labels": ["19.3%","18.9%"],
  "parse_error": false,
  "tool_trace":  {
    "tool":       "ocr_reader_tool",
    "model":      "gemini-2.5-flash-lite",
    "elapsed_ms": 1640.8
  }
}
\end{verbatim}
}

\paragraph{Colour-area fields.} \label{par:mep-colorarea}
The \texttt{color\_area} field is \texttt{null} when gating
conditions are not met (chart type, comparison keyword, or
colour ambiguity checks fail); on this sample the stage was
not triggered. When active it stores:

{\small
\begin{verbatim}
{
  "triggered":            true,
  "breakdown":            {"Series A": 14823, 
                            "Series B": 9102},
  "largest":              "Series A",
  "total_pixels_matched": 23925,
  "low_confidence":       false,
  "color_ambiguity":      false,
  "parse_error":          false
}
\end{verbatim}
}

\section{Development Progression}
\label{sec:appendix_progression}

Table~\ref{tab:progression} shows the iterative development history. It is provided for reproducibility; the main paper reports component contributions rather than version history.

\begin{table}[H]
  \centering
  \scriptsize
  \renewcommand{\arraystretch}{0.9}
  \setlength{\tabcolsep}{2pt}
  \begin{tabular}{p{0.15\columnwidth}p{0.55\columnwidth}p{0.20\columnwidth}}
    \toprule
    \textbf{Version} & \textbf{Key additions} & \textbf{Exact acc.} \\
    \midrule
    Baseline
      & No grounding, default token limits
      & 48.0\% \\
    \addlinespace
    v1
      & Legend grounding, caption injection, token limits raised
      & 50.4\% \\
    \addlinespace
    v2--v3
      & Thinking token exploration; disabled for v4
      & 51.6\% \\
    \addlinespace
    v4
      & Gemini-3 Flash, forced-choice retry, MCQ-aware planner
      & 56.0\% \\
    \addlinespace
    v5
      & Full multi-select MCQ support (planner + vision + verifier)
      & 62.4\% \\
    \addlinespace
    v6--v7
      & Confidence gate with gated revision; silent bug caught
        via MEP trace and repaired
      & 62.8\% \\
    \addlinespace
    v8
      & Deterministic pixel-counting stage
      & 65.1\% \\
    \bottomrule
  \end{tabular}
  \caption{Iterative development history on a fixed development sample ($n \approx 2{,}775$, 25\% of FinMME dataset). The v6--v7 bug was caught by inspecting MEP traces, demonstrating the audit value of per-sample traceability.}
  \label{tab:progression}
\end{table}

\end{document}